\def\year{2022}\relax
\documentclass[letterpaper]{article} 
\usepackage{aaai22}  
\usepackage{times}  
\usepackage{helvet}  
\usepackage{courier}  
\usepackage[hyphens]{url}  
\usepackage{graphicx} 
\usepackage{natbib}  
\usepackage{caption} 
\DeclareCaptionStyle{ruled}{labelfont=normalfont,labelsep=colon,strut=off} 
\usepackage{algorithm}
\usepackage{algorithmic}
\usepackage{amsmath}
\usepackage{hyperref}
\usepackage{makecell}
\usepackage{newfloat}
\usepackage{listings}
\floatstyle{ruled}
\newfloat{listing}{tb}{lst}{}
\floatname{listing}{Listing}
\title{SSAT: A Symmetric Semantic-Aware Transformer Network for Makeup Transfer and Removal}
\author {
    Zhaoyang Sun,\textsuperscript{\rm 1}
    Yaxiong Chen, \textsuperscript{\rm 1} \textsuperscript{\rm 2}
    Shengwu Xiong \textsuperscript{\rm 1} \textsuperscript{\rm 3} \footnote{Corresponding Author}
}
\affiliations {
    \textsuperscript{\rm 1} School of Computer Science and Artificial Intelligence, Wuhan University of Technology, Wuhan, 430070\\
    \textsuperscript{\rm 2} Wuhan University of Technology Chongqing Research Institute, Chongqing, 401122\\
    \textsuperscript{\rm 3} Sanya Science and Education Innovation Park of Wuhan University of Technology, Sanya, 572000\\
    zhaoyangsun0304@outlook.com, chenyaxiong@whut.edu.cn, xiongsw@whut.edu.cn
}

\usepackage{bibentry}

\begin{document}

\maketitle

\begin{abstract}
Makeup transfer is not only to extract the makeup style of the reference image, but also to render the makeup style to the semantic corresponding position of the target image.
However, most existing methods focus on the former and ignore the latter, resulting in a failure to achieve desired results.
To solve the above problems, we propose a unified Symmetric Semantic-Aware Transformer (SSAT) network, which incorporates semantic correspondence learning to realize makeup transfer and removal simultaneously. In SSAT,  a novel Symmetric Semantic Corresponding Feature Transfer (SSCFT) module and a weakly supervised semantic loss are proposed to model and facilitate the establishment of accurate semantic correspondence. In the generation process, the extracted makeup features are spatially distorted by SSCFT to achieve semantic alignment with the target image, then the distorted makeup features are combined with unmodified makeup irrelevant features to produce the final result. Experiments show that our method obtains more visually accurate makeup transfer results, and user study in comparison with other state-of-the-art makeup transfer methods reflects the superiority of our method.
Besides,  we verify the robustness of the proposed method in the difference of expression and pose, object occlusion scenes, and extend it to video makeup transfer.
 Code will be available at \href{https://gitee.com/sunzhaoyang0304/ssat-msp}{SSAT}.
\end{abstract}

\section{Introduction}

\begin{figure}[t]
\centering
\includegraphics[width=0.45\textwidth]{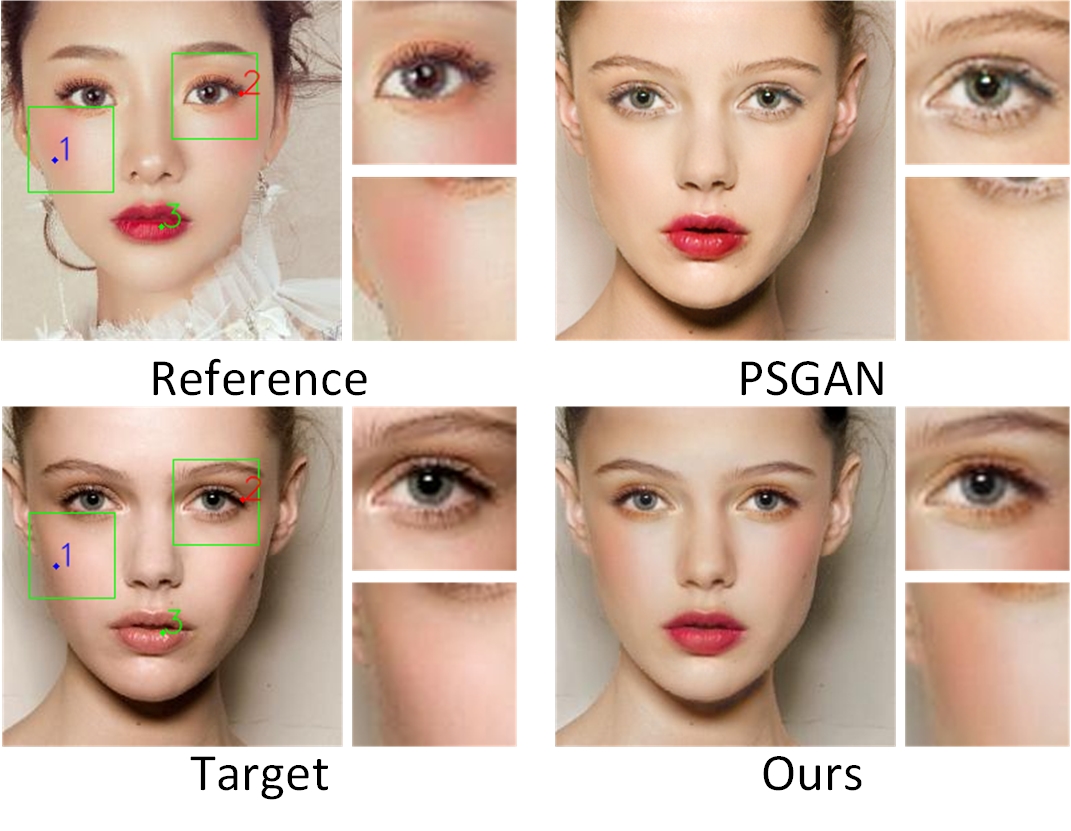} 
\caption{The diagram of  our motivation and PSGAN comparison. The spatial mapping of the same color points between the target image and the reference image is the semantic correspondence expected to be established in this paper.
}
\label{img:motivation}
\end{figure}

\begin{figure}[t]
\centering
\includegraphics[width=0.45\textwidth]{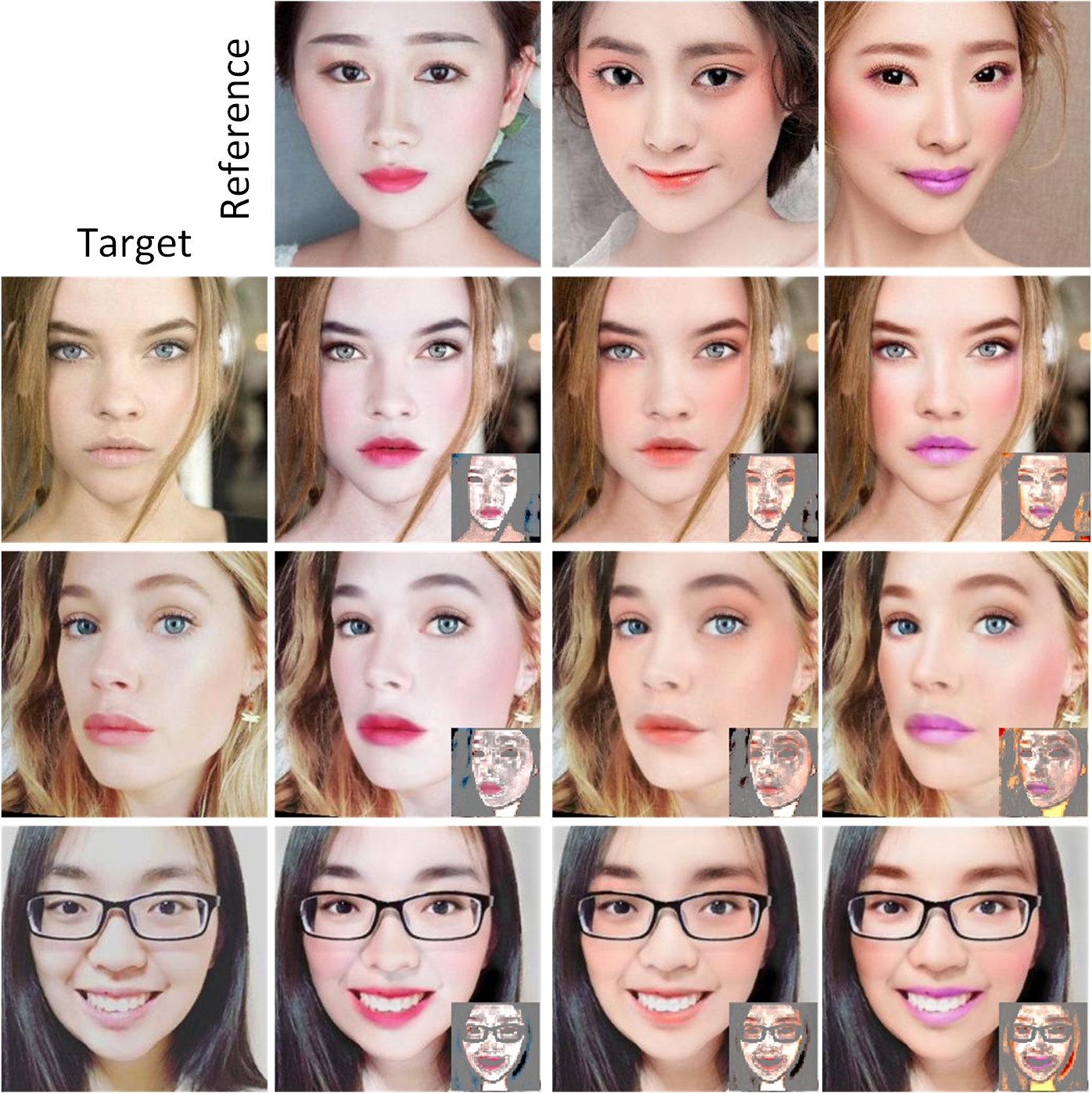} 
\caption{Our SSAT makeup transfer results.
The diversity of makeup style, the difference of facial expression and pose, object occlusion (glasses or hair) are considered in the selected images.
The semantic correspondence is in the lower right corner of the generated result, and the value of each spatial position of it is obtained by the different weighted sum of the reference image.
}
\label{img:transfer_results}
\end{figure}

Makeup transfer aims to transfer the makeup style of any reference image to the target image while preserving the identity of the target person.
With the booming development of the cosmetics market, makeup transfer is widely demanded in many popular beautifying applications and has been received extensive attention in the computer vision and graphics. In the early years, traditional approaches \cite{Guo,Tong,Li} mostly used image gradient editing or physical-based modification to realize makeup transfer. Recently, combining generative adversarial network \cite{GAN} and disentangled representation \cite{DRIT,MUNIT}, many makeup transfer approaches \cite{BeautyGAN,PairedCycleGAN,BeautyGlow,LADN,IPM,SCGAN,CPM} have made significant progress in extracting makeup style and generating realistic makeup results.

However, little attention has been paid to semantic correspondence, which plays an important role in makeup transfer. Consider the actual makeup process following the tutorial: 1) choose cosmetics of the same color (extract makeup style); 2) apply this cosmetics to the semantic corresponding position of the face (semantic correspondence), see Figure \ref{img:motivation}. Ignoring the semantic correspondence would result in the makeup style being transferred to the wrong position, such as lipstick leaking lips. At the same time, inaccurate semantic correspondence will lead to the averaging of makeup colors \cite{PSGAN}, so that the resulting makeup style is visually different from the reference makeup. In addition, the robustness of expression and pose and the robustness of object occlusion are greatly reduced.

The motivation of this paper is to establish accurate semantic correspondence between the target and the reference facial images to improve the quality of makeup transfer.  Note that the dense semantic correspondence of the makeup regions is established, not the sparse semantic correspondence shown in the Figure\ref{img:motivation}.
For this purpose,  a  Symmetric Semantic Corresponding Feature Transfer (SSCFT) module is proposed to  models the spatial position mapping between different images.
Next there is no effective supervision for semantic correspondence, so the face parsing is introduced and a weakly supervised semantic loss is proposed. The experiments show that the semantic correspondence established by our method  is adaptive, which ignores the semantic correspondence  of makeup irrelevant regions (hair, background, glasses, inside eyes, inside mouth) and focuses on the makeup regions, see Figure \ref{img:transfer_results}.

\begin{figure*}[t]
\centering
\includegraphics[width=0.8\textwidth]{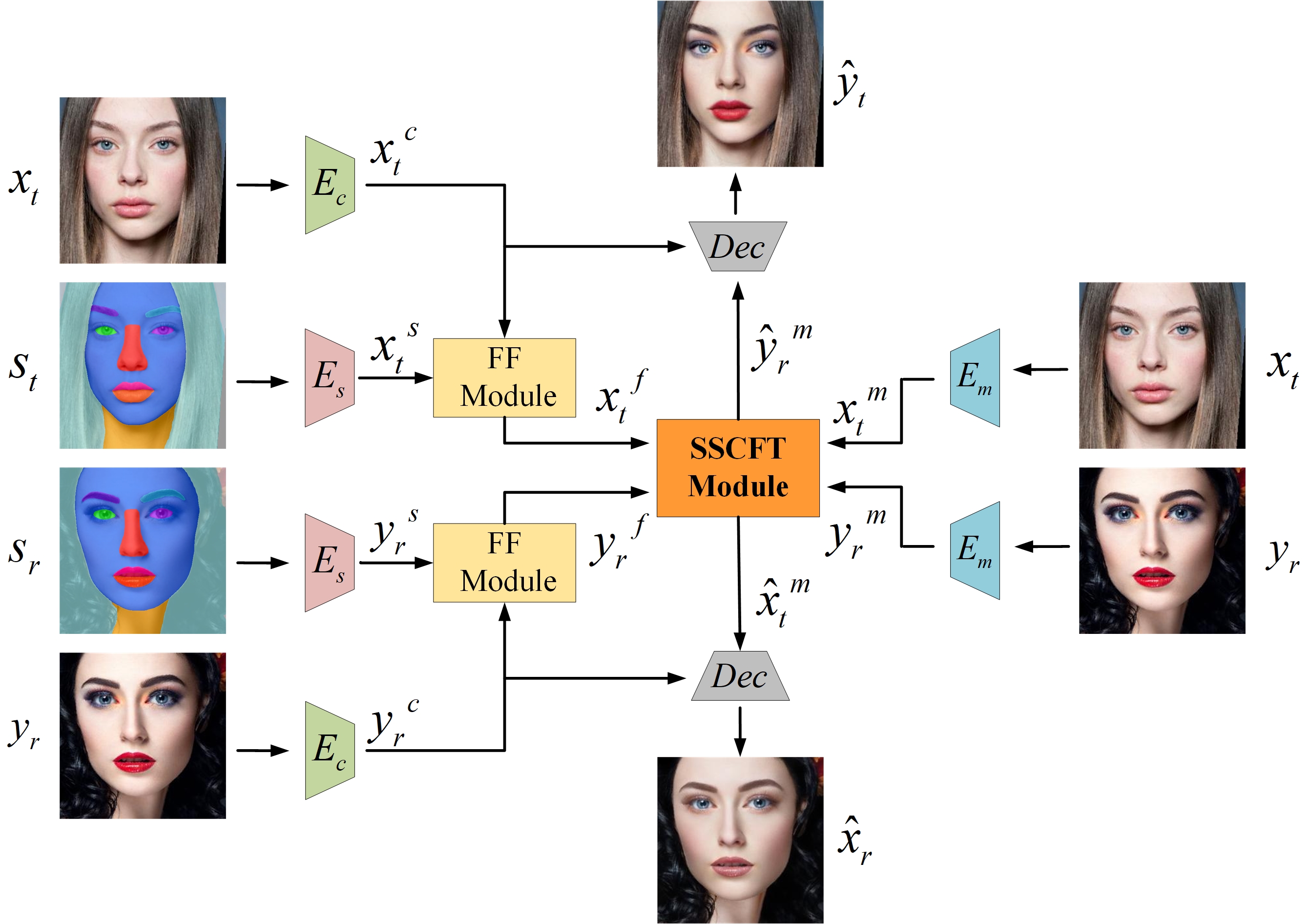} 
\caption{The proposed Symmetric Semantic-Aware Transformer (SSAT) network. The process goes through the following steps: 1) Content encoder $E_{c}$ and makeup encoder $E_{m}$ decompose target image $x_{t}$ and reference image $y_{r}$ respectively, $x_{t}^{c}=E_{c}(x_{t}), x_{t}^{m}=E_{m}(x_{t}), y_{r}^{c}=E_{c}(y_{r}),  y_{r}^{m}=E_{m}(y_{r})$. Meanwhile, face parsing $s_{t}, s_{r}$ is introduced and semantic features are extracted by using semantic encoders $E_{s}$: $x_{t}^{s}=E_{s}(s_{t}),  y_{r}^{s}=E_{s}(s_{r})$.
2) The Feature Fusion (FF) module fuses content features and semantic features to obtain richer features for semantic correspondence, $x_{t}^{f}=FF(x_{t}^{c},x_{t}^{s}), y_{r}^{f}=FF(y_{r}^{c},y_{r}^{s})$.
3) The Symmetric Semantic Corresponding Feature Transfer (SSCFT) module distorts makeup features spatially according to the semantic correspondence established by $x_{t}^{f}$ and $y_{r}^{f}$, and outputs $\hat{y}_{r}^{m}, \hat{x}_{t}^{m}=SSCFT(x_{t}^{f},y_{r}^{f},x_{t}^{m},y_{r}^{m})$.
4) Distorted makeup features $\hat{y}_{r}^{m}$ of the reference image are embedded in the content features $x_{t}^{c}$ of the target image to generate makeup transfer result $\hat{y}_{t}= Dec(x_{t}^{c}, \hat{y}_{r}^{m})$. Similarly, the makeup removal result $\hat{x}_{r}=  Dec(y_{r}^{c}, \hat{x}_{t}^{m})$.
}
\label{img:network}
\end{figure*}

To sum up, we propose a Symmetric Semantic-Aware Transformer network (SSAT) whose framework and process are shown in Figure \ref{img:network}.
Compared with PSGAN \cite{PSGAN}, there are several obvious differences: 1) The proposed method introduces the face parsing, instead of the feature points which sometimes has large error in side face. 2) Learnable network features instead of hand-designed features to establish semantic correspondence. 3) Using the symmetry of semantic correspondence, our method generates makeup transfer and removal results simultaneously, while PSGAN can only realize makeup transfer.
The main contributions of this paper are summarized as follows:
\begin{itemize}
  \item We propose a novel Symmetric Semantic-Aware Transformer (SSAT) network for makeup transfer and removal. Experiments verify the effectiveness of the transferring strategy in the real-world environment and generated results are of higher quality than state-of-the-art methods both qualitatively and quantitatively.
  \item A novel SSCFT module and a semantic loss are proposed to establishing accurate semantic correspondence, which significantly improves the quality of makeup transfer.
  \item Combined with face parsing, our method is more flexible and could realize partial makeup transfer. Meanwhile, we verify the robustness of the proposed method in the difference of expression and pose, object occlusion (glasses or hair) scenes and extend it to video makeup transfer.
\end{itemize}

\section{Related work}
\subsection{Makeup Transfer} In recent years, makeup transfer has been extensively studied. BeautyGAN \cite{BeautyGAN} addressed the makeup transfer and removal task
by incorporating both global domain adaptation loss and local instance-level makeup loss in an dual input/output GAN. PairedCycleGAN \cite{PairedCycleGAN} extended the CycleGAN \cite{CycleGAN} to asymmetric networks to enable transferring specific makeup style.
LADN \cite{LADN} and CPM \cite{CPM} focused on the complex/dramatic makeup styles transfer.
Introducing facial feature points, PSGAN \cite{PSGAN,PSGAN++} proposed a pose and expression robust spatial-aware GAN  for makeup transfer.
Recently, SOGAN \cite{SOGAN} explored the shadow and occlusion robust and \cite{Beihang} realized makeup transfer from the perspective of face attribute editing.
Inspired by StyleGAN \cite{StyleGAN}, SCGAN \cite{SCGAN} proposed a style-based controllable GAN model. Unlike the above methods, the core idea of this paper is to establish accurate semantic correspondence to improve the quality of makeup transfer.

\subsection{Semantic Correspondence}
In recent years, CNN-based features \cite{VGG,Alex} have been proved to be a powerful tool to express high-level semantics.
Recently, \cite{VisualAttribute,CoCosNet} proposed a technique for visual attribute transfer across images using semantic correspondence. The exemplar-based colorization methods \cite{DeepColorization,SketchColorization} calculated the semantic correspondence between the target image and the exemplar image, then transferred the color with the closest semantic similarity to the target image.
Inspired by the recent exemplar-based image colorization \cite{DeepColorization,SketchColorization}, our work is expected to transfer the makeup style with the closest semantic similarity to the target image.

\section{Our approach: SSAT}

\subsection{Formulation}
Our goal is to transfer the makeup style from an arbitrary makeup reference  image to a non-makeup target  image.
Here, $X \subset \mathcal{R}^{H \times W \times 3} $ refers to non-makeup image domain, $Y \subset \mathcal{R}^{H \times W \times 3} $ refers to makeup image domain.
Given a target image $x_{t}\in X$ and a reference image $y_{r} \in Y$, the goal of makeup transfer is learning a mapping function:
$ \Phi: x_{t} , y_{r} \rightarrow \hat{y}_{t} $, where $ \hat{y}_{t} \in Y $ has the makeup style with $ y_{r}$ while preserving the identity of $x_{t}$.
For makeup removal, it is assumed that the non-makeup image is a special case of the makeup image \cite{LocalMakeup}, which unifies makeup transfer and makeup removal.
Therefore, the goal of makeup removal is learning a mapping function:
$ \Phi: y_{r}, x_{t} \rightarrow \hat{x}_{r} $, where $ \hat{x}_{r} \in X $ has the makeup style with $ x_{t}$ while preserving the identity of $y_{r}$.
In this paper, the only difference between makeup transfer and removal is whether the reference image is a non-makeup image or a makeup image.

\subsection{SSAT}
The overall framework of  Symmetric Semantic-Aware Transformer network (SSAT) is shown in Figure \ref{img:network}, which consists of three encoders, a FF module, a SSCFT module and a decoder $Dec$. Next, the function of each module will be introduced in detail.

\subsubsection{Encoders}
The one main problem of makeup transfer stems from the difficulty of extracting the makeup latent features, which are required to be disentangled from other makeup irrelevant features. Here, this problem is referred to as content-style separation \cite{MUNIT,DRIT}.
So  a content encoder $E_{c}$ and a makeup encoder $E_{m}$ are designed to extract content features and makeup features respectively:
\begin{equation}\label{equ1}
  x_{t}^{c}=E_{c}(x_{t}),  x_{t}^{m}=E_{m}(x_{t})
\end{equation}
\begin{equation}\label{equ2}
  y_{r}^{c}=E_{c}(y_{r}),  y_{r}^{m}=E_{m}(y_{r})
\end{equation}
Experiments have found that it is difficult to establish accurate semantic correspondences only with content features. Therefore, face parsing \cite{BiSeNet} is introduced and semantic features are extracted by using semantic encoders $E_{s}$:
\begin{equation}\label{equ3}
  x_{t}^{s}=E_{s}(s_{t}),  y_{r}^{s}=E_{s}(s_{r})
\end{equation}
where $s_{t} \in \mathcal{R}^{H \times W \times L}$ and $s_{r} \in \mathcal{R}^{H \times W \times L}$ refer to binary face parsing of the target image and the reference image, respectively.  $L$ is the number of different semantic regions, which is set 18 in our experiments. Next, content features and semantic features will cooperate to establish semantic correspondence.

\subsubsection{FF}
\begin{figure}[t]
\centering
\includegraphics[width=0.45\textwidth]{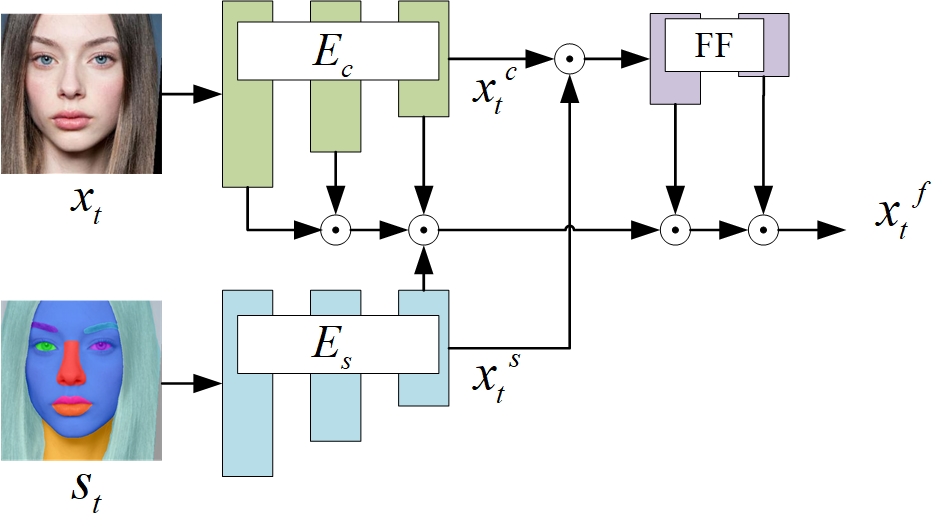} 
\caption{The illustration of feature fusion (FF) module. FF fuses content features and semantic features to obtain richer features for semantic correspondence. $\odot$ refers to the channel-wise concatenation operator. }
\label{img:FF}
\end{figure}
In the FF module, content features and semantic features are fused  to obtain richer features for feature matching, see Figure \ref{img:FF}.
Take the target image as an example, after obtaining content features and semantic features, they are connected along the channel dimensions and fed into the FF module. FF consists of $N$ convolutional layers and outputs $N$ high-level features $(f^{1},f^{2},\cdots,f^{N})$. Besides, the low-level features extracted by $E_{c}$ and $E_{s}$ are also combined.  $E_{c}$ and $E_{s}$ consist of $M$ convolutional layers, producing $M$ low-level features $(C^{1},C^{2},\cdots,C^{M})$ and $(S^{1},S^{2},\cdots,S^{M})$, where $C^{M}=x_{t}^{c}, S^{M}=x_{t}^{s} $. Each layer of content features is selected, but the last layer of semantic features is selected. Now, we downsample each of $C^{m}$ and  upsample each of $f^{n}$ to match the spatial size $S^{M}$, forming the final fusion features $x_{t}^{f}$:
\begin{equation}\label{equ4}
\begin{aligned}
  x_{t}^{f}=[& \phi(C^{1}); \phi(C^{2}); \cdots; C^{M}; S^{M}; \\
  & \varphi(f^{1});\varphi(f^{2});\cdots;\varphi(f^{N})]
\end{aligned}
\end{equation}
where $\phi$ denotes a spatially downsampling function of an input $C^{m}$ of different size to the size of  $C^{M}$ or $S^{M}$. Similarly, $\varphi$ denotes a spatially upsampling function. "$;$" denotes the channel-wise concatenation operator. In this manner, the output fusion feature $x_{t}^{f}$ combines the low-level and high-level features,  while ensuring the accuracy of semantic correspondence. With the same operation, we obtain the fusion features $y_{r}^{f}$ of the reference image.

\subsubsection{SSCFT}
\begin{figure}[t]
\centering
\includegraphics[width=0.45\textwidth]{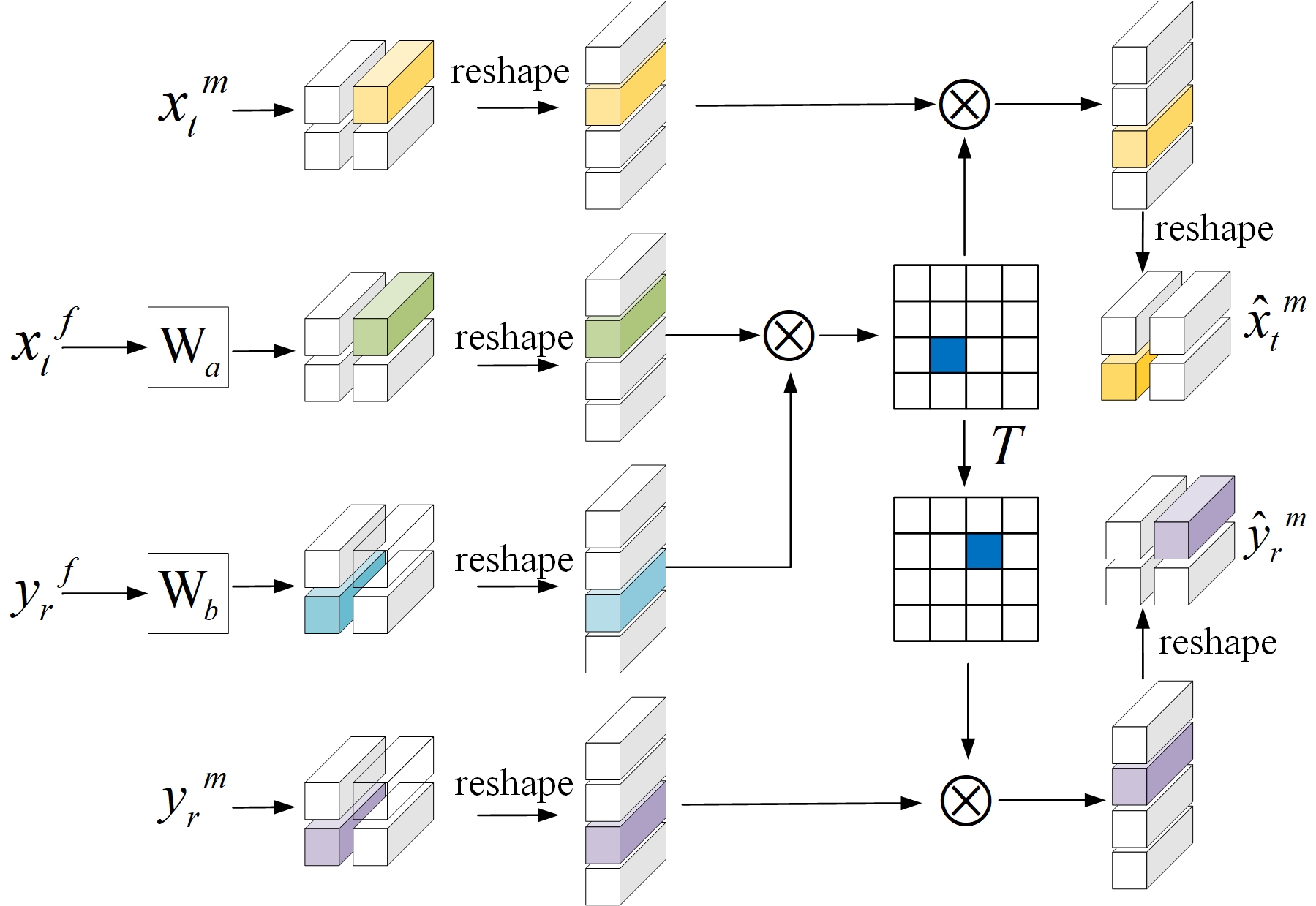} 
\caption{The illustration of symmetric semantic corresponding feature transfer (SSCFT) module. $W_{a}, W_{b}$ refer to the linear transformation matrix into $ x_{t}^{f}$ and $ y_{r}^{f}$, respectively. $\otimes$ refers to the matrix multiplication operator. $T$ refers to the matrix transpose operator. After this module, the output distorted makeup features $\hat{x}_{t}^{m}, \hat{y}_{r}^{m}$ are semantic aligned with $y_{r}^{c},x_{t}^{c}$ in semantic.}
\label{img:SSCFT}
\end{figure}
The SSCFT module is inspired by colorization \cite{DeepColorization} and combines the symmetry of semantic correspondence. In makeup transfer task, the semantic correspondence is a one-to-one mapping relationship. That is, point A corresponds to point B. In turn, point B also corresponds to point A. And the symmetry of this semantic relationship can be applied cleverly to the anti-task of makeup transfer, makeup removal. See Figure \ref{img:SSCFT} for the framework  of the proposed SSCFT module.

Firstly, we reshape $ x_{t}^{f}$ into $ \hat{x}_{t}^{f}=[x_{1},x_{2},\cdots,x_{hw}] \in \mathcal{R}^{d \times hw}$, where $x_{i} \in \mathcal{R}^{d} $ indicates feature variables of the $i^{th}$ location of $ \hat{x}_{t}^{f}$. Then we obtain  channel-wise centralized features $\hat{x}_{i}$  and $\hat{y}_{j}$ to make the learning more stable \cite{DeepColorization},  where $\hat{x}_{i}=x_{i}-mean(x_{i}),\hat{y}_{j}=x_{y}-mean(y_{j})$.  Given  $\hat{x}_{i}$ and $\hat{y}_{j}$, SSCFT computes a semantic correlation matrix $\mathcal{A} \in \mathcal{R}^{hw \times hw}$, whose element $a_{i,j}$ is computed by the scaled dot product:
\begin{equation}\label{equ5}
  a_{i,j}= \frac{\hat{x}_{i}^{T} \cdot \hat{y}_{j}}{\|\hat{x}_{i}\| \|\hat{y}_{j}\|}
\end{equation}
After that, we distort the reference makeup features $ y_{r}^{m}$  towards $ x_{t}^{c}$  according to the correlation matrix  $\mathcal{A}$. The weighted sum of
$\hat{y}_{r}^{m}$ is calculated to approximate the makeup sampling from  $ y_{r}^{m}$ :
\begin{equation}\label{equ6}
  \hat{y}_{r}^{m}(i)= \sum_{j} softmax_{j} (a(i,j) \cdot \sigma)\cdot y_{r}^{m}(j)
\end{equation}
where $\sigma$ controls the sharpness of the softmax and we set its default value as 100. Now the distorted makeup features $\hat{y}_{r}^{m}$ of reference image are aligned with the content features $x_{t}^{c}$ of target image in semantic. In the same way, we obtain the distorted makeup features $\hat{x}_{t}^{m}$, which aligns with the content features $y_{r}^{c}$. Note that this step makes our method robust to  expression, pose, object occlusion and produce more accurate makeup transfer results.
\subsubsection{Dec}
Finally, we employ the spatially-adaptive denormalization (SPADE) block \cite{SPADE} to project the distorted makeup styles $\hat{y}_{r}^{m}, \hat{x}_{t}^{m} $ to content features $x_{t}^{c}, y_{r}^{c} $ for makeup transfer and removal.
\begin{equation}\label{equ7}
  \hat{y}_{t}= Dec(x_{t}^{c}, \hat{y}_{r}^{m})
\end{equation}
\begin{equation}\label{equ8}
  \hat{x}_{r}=  Dec(y_{r}^{c}, \hat{x}_{t}^{m})
\end{equation}
where $\hat{y}_{t}$ is the makeup transfer result and $\hat{x}_{r}$ is the makeup removal result.

\subsection{Objective}
In total, there are four loss functions used for network SSAT end-to-end training. The overall loss is as follows:
\begin{equation}\label{equ9}
\begin{aligned}
    L_{overall}=   &\lambda_{sem}L_{sem}+\lambda_{makeup}L_{makeup} +\\
  &  \lambda_{rec}L_{rec}+\lambda_{adv}L_{adv}
\end{aligned}
\end{equation}

\subsubsection{Semantic Loss} A weakly supervise semantic loss is proposed to establish semantic correspondence. The idea is that semantic correspondence should only exist between semantic regions of the same class:
\begin{equation}\label{equ_semantic}
    L_{sem}= \|s_{t}- \hat{s}_{r}\|_{1}+\|s_{r}- \hat{s}_{t}\|_{1}
\end{equation}
where $s_{t} \in \mathcal{R}^{H \times W \times L}$ and $s_{r} \in \mathcal{R}^{H \times W \times L}$  refer to binary
face parsing, which only have the values 0 and 1. $L$ is the number of semantic classes,
$ \hat{s}_{r}(i)= \sum_{j} softmax_{j} (a(i,j) \cdot \sigma)\cdot s_{r}(j)$, $ \hat{s}_{t}(i)= \sum_{j} softmax_{j} (a(i,j) \cdot \sigma)\cdot s_{t}(j)$, and $\|\cdot\|_{1}$ refers to $L_{1}$ loss. For dense semantic correspondence, the $L_{sem}$ is only a crude region constraint, but experiments show that it plays an important role.

\begin{figure}[t]
\centering
\includegraphics[width=0.45\textwidth]{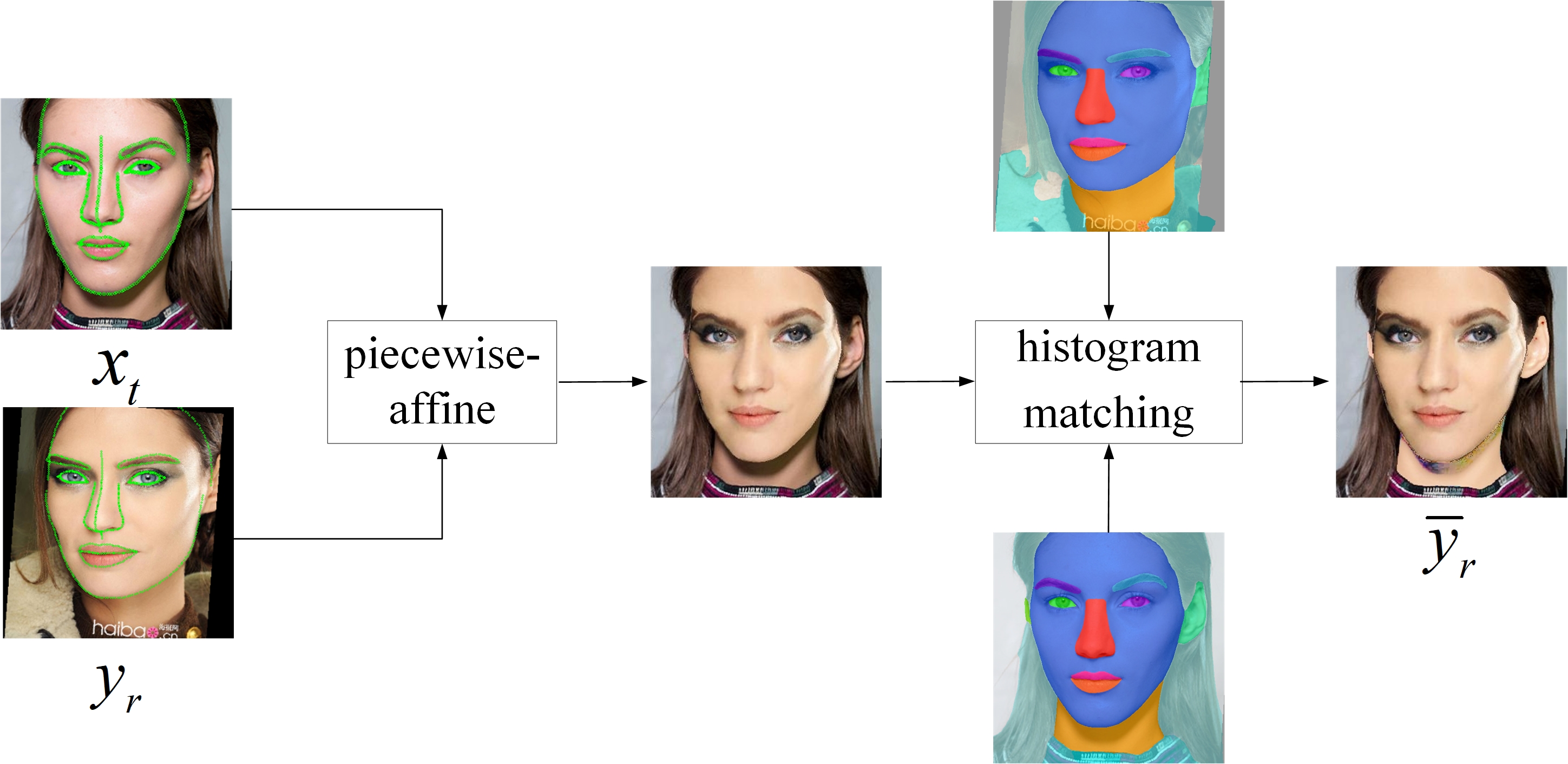} 
\caption{The process of generating pseudo-paired data.}
\label{img:groundtrue}
\end{figure}

\subsubsection{Makeup Loss} Inspired by \cite{PairedCycleGAN}, we generate pseudo pairs of data $\bar{x}_{r}$ and $\bar{y}_{t}$ according to face feature points to train the network, see Figure \ref{img:groundtrue}. The process is described in detail in the supplementary materials. Here, we introduce the SPL loss \cite{SPL} instead of the $L_{1}$ loss to guide the makeup transfer and removal:
\begin{equation}\label{equ_makeup}
    L_{makeup}= SPL(\hat{y}_{t}, x_{t}, \bar{y}_{t}) + SPL(\hat{x}_{r}, y_{r}, \bar{x}_{r})
\end{equation}
Take the makeup transfer as an example, SPL constrains the gradient consistency between $\hat{y}_{t}$ and $x_{t}$ to ensure the identity of the target image, and restricts the color consistency between $\hat{y}_{t}$ and $\bar{y}_{t}$ to guide the makeup transfer. Note that $L_{sem}$ constrains the corresponding region and does not penalize one-to-many mappings of the same class. While $L_{makeup}$ guides makeup transfer, it implicitly constrains the one-to-one mapping in the makeup regions.

\subsubsection{Reconstruction loss} We feed the $x_{t}^{c}$ and $x_{t}^{m}$ into $Dec$ to generate $x_{t}^{self}$,  and $y_{r}^{c}$ and $y_{r}^{m}$ into $Dec$ to generate $y_{r}^{self}$, which should be identical to $x_{t}$ and $y_{r}$.
Here, we introduce Cycle loss \cite{CycleGAN} to ensure that the image does not lose information during the decoupling process of makeup features and content features.
So we feed the makeup removal result $\hat{x}_{r}$ and makeup transfer $\hat{y}_{t}$ result into SSAT again to obtain the  $x_{t}^{cycle}$ and $y_{r}^{cycle}$, which should also be identical to $x_{t}$ and $y_{r}$. We use L1 loss to encourage such reconstruction consistency:

\begin{equation}\label{equ_rec}
\begin{aligned}
    L_{rec}= & \|x_{t}- x_{t}^{self}\|_{1}+\|y_{r}- y_{r}^{self}\|_{1}+ \\
    & \|x_{t}- x_{t}^{cycle}\|_{1}+\|y_{r}- y_{r}^{cycle}\|_{1}
\end{aligned}
\end{equation}

\subsubsection{Adversarial loss}
Two discriminators $D_{X}$ and $D_{Y}$ are introduced for the non-makeup domain and makeup domain, which try to discriminate between real samples and generated images and help the generator synthesize realistic outputs. The least square loss \cite{lsgan} is used for steady training:
\begin{equation}\label{equ_adv}
\begin{aligned}
    L_{adv}=  &\mathbf{E}_{x_{t}}[(D_{X}(x_{t}))^{2}]+\mathbf{E}_{\hat{x}_{r}}[(1-D_{X}(\hat{x}_{r}))^{2}]+\\  &\mathbf{E}_{y_{r}}[(D_{Y}(y_{r}))^{2}]+\mathbf{E}_{\hat{y}_{t}}[(1-D_{Y}(\hat{y}_{t}))^{2}]
\end{aligned}
\end{equation}

\section{Experiment}

\subsection{Dataset and Implementation Details}

For the dataset, we randomly selected 300 non-makeup images as target images and 300 makeup images as reference images from the Makeup Transfer dataset \cite{BeautyGAN}. Using the  proposed generation method, a total of 180,000 pairs of pseudo-paired data are generated for makeup transfer and removal.
During the training, all trainable parameters are initialized normally, and the Adam optimizer with $\beta_{1}=0.5,\beta_{2}=0.999$ is employed for training. We set $\lambda_{sem}=1,\lambda_{makeup}=1,\lambda_{rec}=1,\lambda_{adv}=1$ for balanceing the different loss functions.
The SSAT is implemented by MindSpore \footnote{Mindspore. https://www.mindspore.cn/}. And the model is trained for 300,000 iterations on one single  Nvidia 2080Ti GPU.
The learning rate is fixed as 0.0002 during the first 150,000 iterations and linearly decays to 0 over the next 150,000 iterations. The batchsize is set 1.
See the supplementary materials for the specific network structure parameters.
%

\subsection{Ablation Studies}
\begin{figure}[t]
\centering
\includegraphics[width=0.45\textwidth]{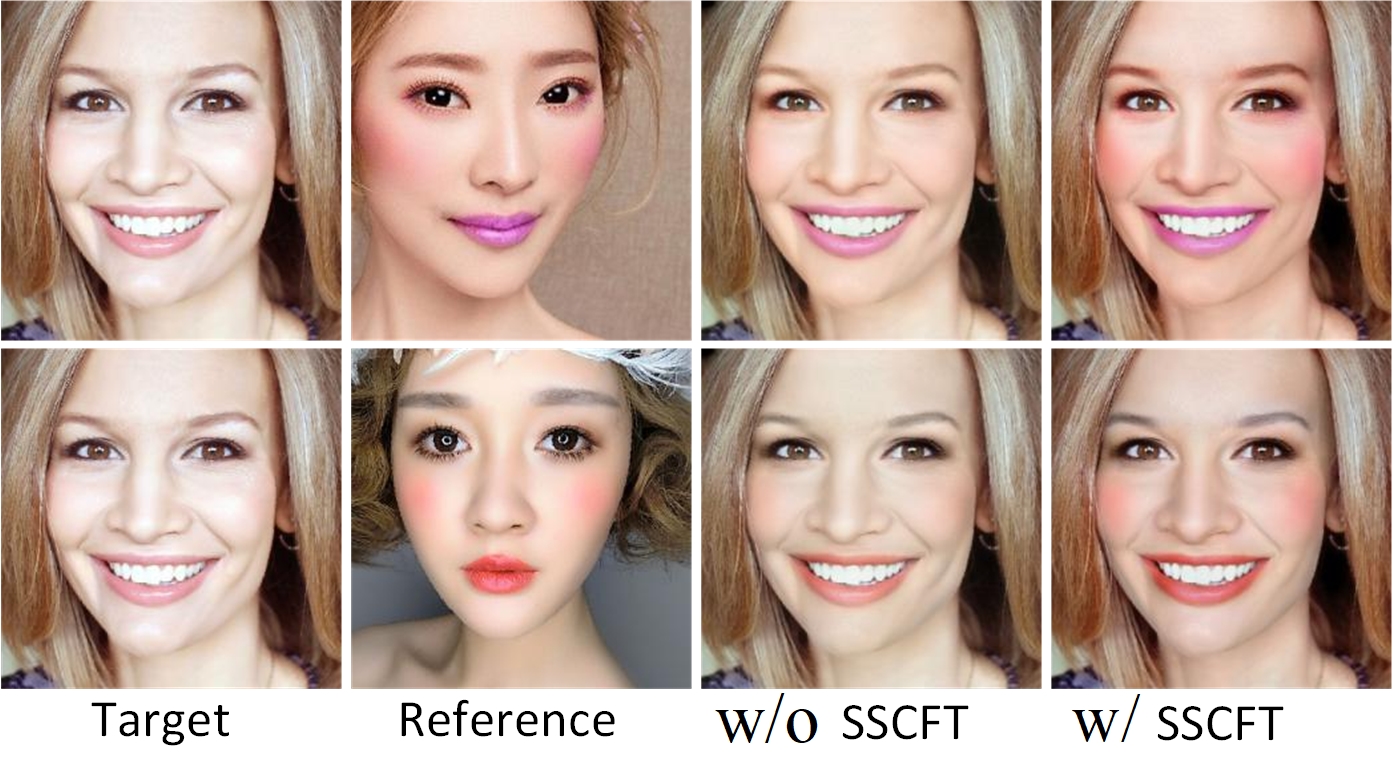} 
\caption{Ablation study of SSCFT module.
}
\label{img:ablation_SSCFT}
\end{figure}
\subsubsection{Effects of SSCFT} The motivation of this paper is to establish semantic correspondence to improve the quality of makeup transfer.
In order to verify our idea, we remove the SSCFT module to analyze the effects of the role of semantic correspondence in makeup transfer.
In this case, we directly skip the SSCFT module and input the features into the decoder.
The comparison results are shown in the Figure \ref{img:ablation_SSCFT}.
Without SSCFT, the resulting makeup style is significantly lighter than the reference makeup.
With the addition of SSCFT, the result is more similar visually to the reference makeup  , especially eye shadow and blush.

\subsubsection{Effects of $L_{sem}$}
\begin{figure}[t]
\centering
\includegraphics[width=0.4\textwidth]{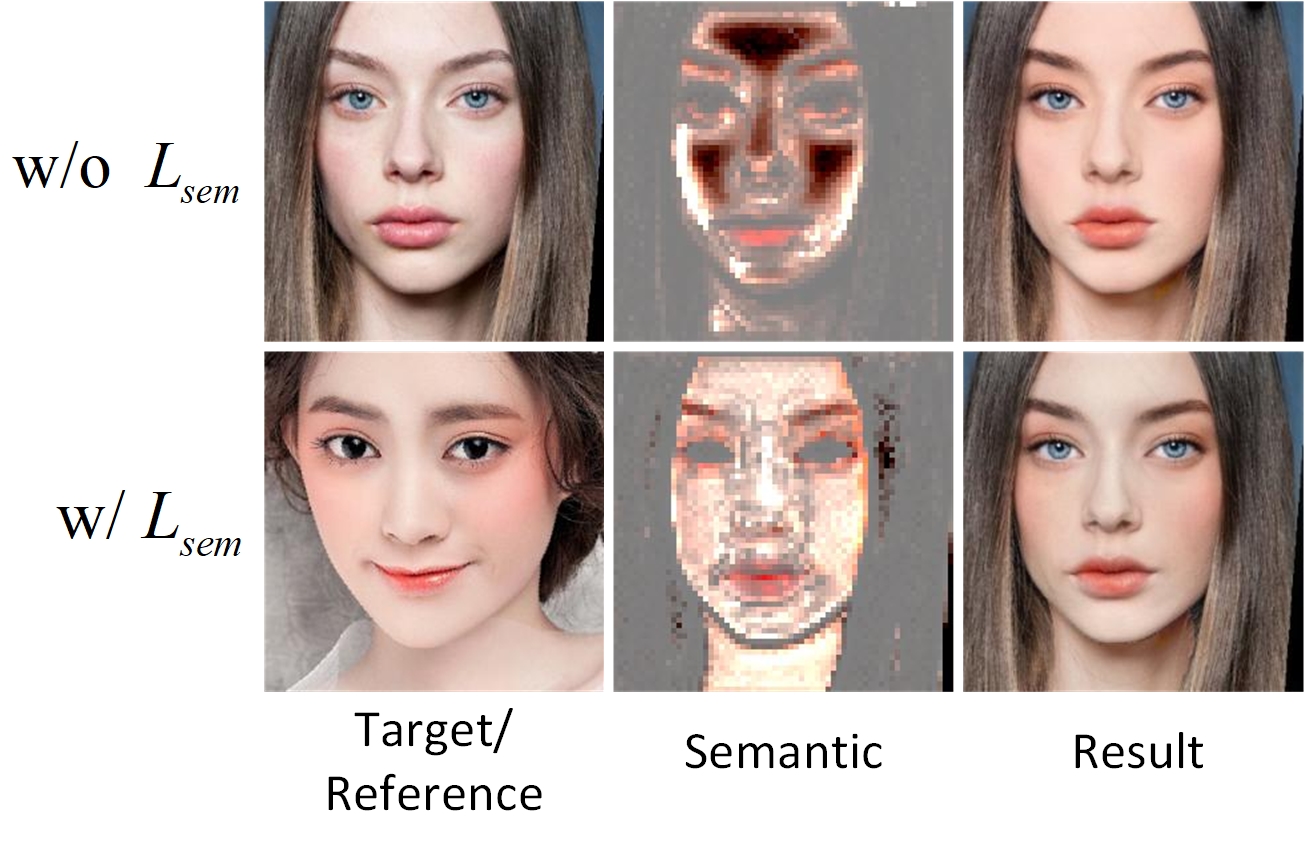} 
\caption{Ablation study of $L_{sem}$.
}
\label{img:ablation_sem_loss}
\end{figure}
The lack of effective supervision hinders the establishment of semantic correspondence. To solve this problem, we introduce face parsing and propose a weakly supervised semantic loss $L_{sem}$.
In order to verify their effect, we remove $L_{sem}$ and the face parsing, use the remaining loss to train the network.
The comparison of semantic and makeup transfer results is shown in the Figure \ref{img:ablation_sem_loss}.
Compared without $L_{sem}$, the semantic result using $L_{sem}$ is more accurate and the boundary between makeup area and non-makeup area is also clearer.
For the makeup transfer result, the blush is mapped to the wrong semantic position, spreading over the entire face in the result without using $L_{sem}$.

\subsection{Comparisons to Baselines}
To verify the superiority of our makeup transfer strategy, we choose three state-of-the-art  makeup transfer approaches, BeautyGAN \cite{BeautyGAN}, PSGAN \cite{PSGAN}, SCGAN \cite{SCGAN}, as our comparison benchmark. We skip some baselines such as LADN \cite{LADN} and CPM \cite{CPM}, because they focus on the complex/dramatic makeup styles transfer.
\subsubsection{Qualitative Comparison}
\begin{figure*}[t]
\centering
\includegraphics[width=0.75\textwidth]{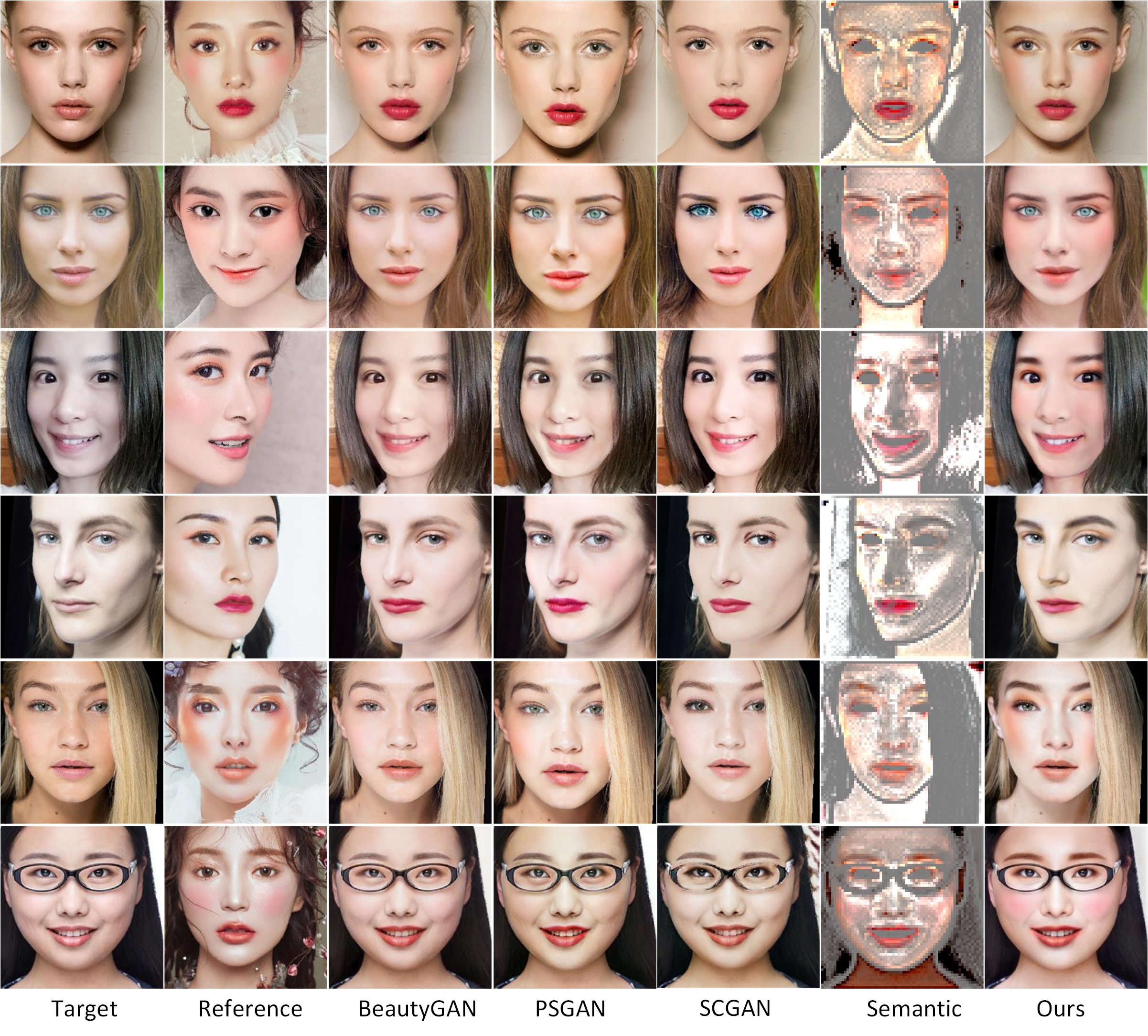} 
\caption{Comparison with state-of-the-art methods. The first two rows: frontal face, the middle two rows:  different expressions and poses, the last two rows: object occlusion (glasses or hair).
}
\label{img:qualitative}
\end{figure*}

Here, three scenes that often appear in real life are selected for comparison, frontal face, different expressions and poses, and object occlusion. The qualitative comparison has been shown in Figure \ref{img:qualitative}.
BeautyGAN produces a realistic result, but fails to transfer eye shadow and blush.
Although PSGAN designs semantic correspondence modules, its accuracy is limited by hand-designed features. The imprecise semantic correspondence results in a lighter eye and blush makeup style, which is visually different from the reference makeup.
Similar to BeautyGAN, the transfer of eye makeup and blush fails in SCGAN. Meanwhile, the results of SCGAN are too smooth, and slightly change makeup irrelevant information, such as the glasses in the sixth row.
 On the contrary,  the makeup style of our results is highly similar to the reference makeup, whether it is lipstick, eye shadow, blush. The semantic results explain why our method has a better makeup transfer effect and why our method is robust to expression, pose and  object occlusion.

\subsubsection{Quantitative Comparison}


\begin{table}[t]
\centering
\begin{tabular}{c|c|c|c|c}
    \hline
    Methods & Rank 1 & Rank 2& Rank 3& Rank 4 \\
    \hline
    \hline
    BeautyGAN & 3.0\%&3.9\%&34.5\%&58.6\% \\
    \hline
    PSGAN &   11.3\%&73.2\%&11.3\%&4.1\%\\
     \hline
    SCGAN &   3.6\%&15.3\%&48.0\%&33.2\% \\
    \hline
    SSAT &   82.1\%&7.6\%&6.2\%&4.1\%\\
    \hline
\end{tabular}
\caption{User Study.}
\label{table2}
\end{table}

\begin{figure}[t]
\centering
\includegraphics[width=0.45\textwidth]{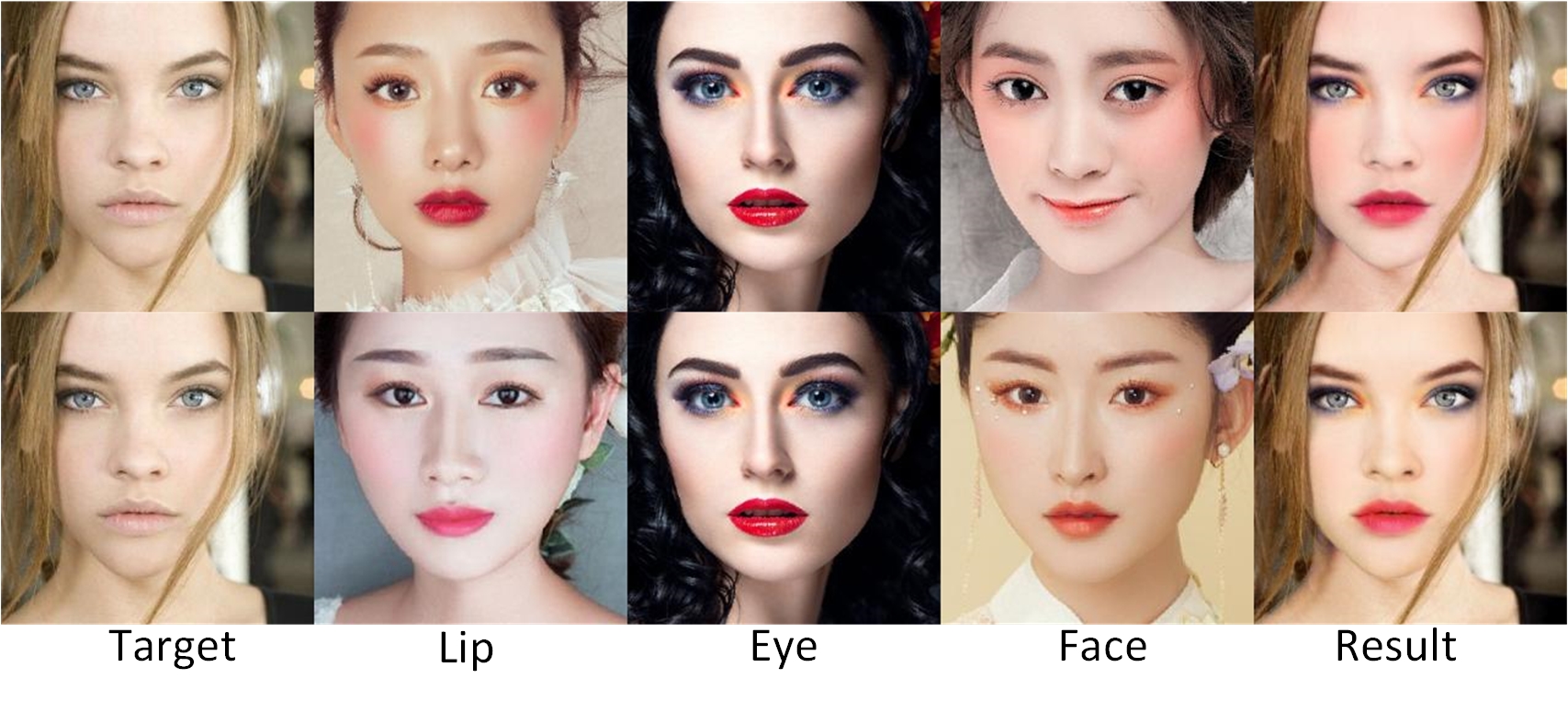} 
\caption{The partial makeup transfer results. The last column is the partial makeup transfer results, which receive personal identity from the first column, the lips style
from the second column, the eyes style from the third column and the face style from the fourth column.
}
\label{img:local_makeup_transfer}
\end{figure}


\begin{figure}[t]
\centering
\includegraphics[width=0.4\textwidth]{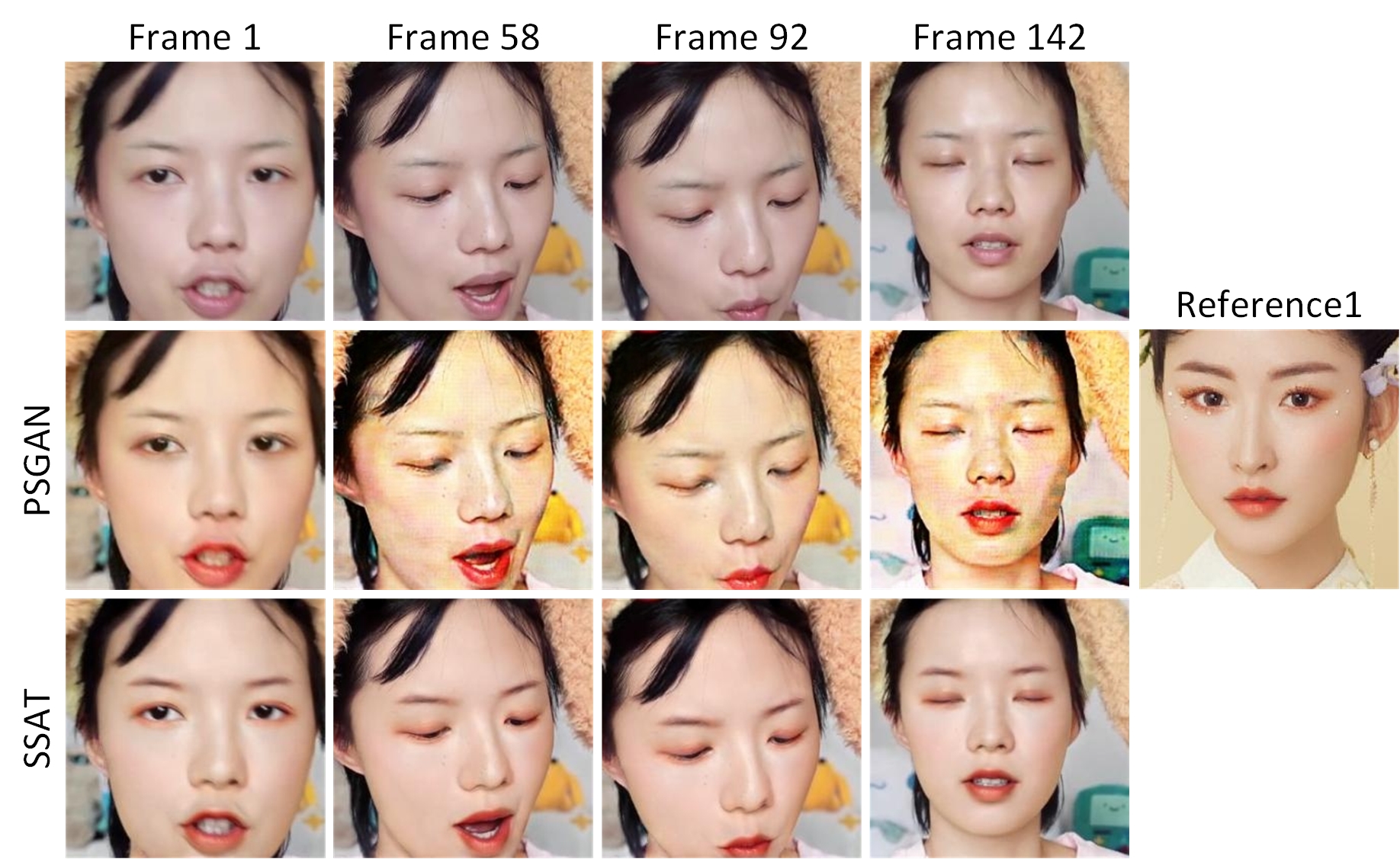} 
\caption{The video makeup transfer results.
}
\label{img:video_makeup_transfer}
\end{figure}

How quantitative evaluation makeup transfer is still a field that needs to explore. Here, we conduct user study to compare different methods quantitatively.
We randomly generate 30 results of makeup transfer using four methods respectively.
45 volunteers are asked to rank the results  based on the realism and the similarity of makeup styles.
To be fair, the results in each selection are also randomly arranged.
As shown in the Table \ref{table2}, our SSAT outperforms other methods by a large margin. Our method achieves the highest selection rate of 82.1\% in Rank 1.

\subsection{Partial Makeup Transfer}
Partial makeup transfer refers to transfer partial makeup of the reference image.
The distorted makeup features extracted by our method is accurately distributed according to the  spatial semantic position of the target image, making it possible  to integrate the partial makeup from different reference images, as shown in the Figure \ref{img:local_makeup_transfer}.

\begin{equation}\label{equ14}
  \hat{y}_{t}^{part}= Dec(x_{t}^{c},\sum_{i} (\hat{y}_{r}^{m_{i}} \cdot Mask_{i}))
\end{equation}
where $i \in \{Lip, Eye, Face\}$ in our experiment, $\hat{y}_{r}^{m_{i}}$ means the distorted makeup features extracted are from different reference images, $Mask_{i}$ represents a binary mask related to the makeup area.
%

\subsection{Video Makeup Transfer}
Video makeup transfer is a very challenging task, which has high requirements for the quality of generated images and the accuracy of semantic correspondence.
We download a video from the Internet and decompose it frame by frame, then apply the SSAT method, and finally integrate the resulting images into a video.
We chose PSGAN as the comparison baseline, because other methods don't consider semantic correspondence.
See Figure \ref{img:video_makeup_transfer}, the results produced by PSGAN are visually different from the reference makeup and cause flickering and discontinuity.
In contrast, Our SSAT achieves smooth and accurate video makeup transfer results.

\section{Conclusion}
Different from other methods, we focus on semantic correspondence learning, propose the SSCFT module and a semantic loss, then integrate them into one Symmetric Semantic-Aware Transformer network (SSAT) for makeup transfer and removal. The experiment verified that semantic correspondence significantly improved the quality of makeup transfer visually as expected. The comparison with other methods demonstrates that our method achieves  state-of-the-art makeup transfer results. In addition, benefits from precise semantic correspondence, our method is robust to the difference of expression and pose, object occlusion and  can achieve partial makeup transfer. Moreover, we extend SSAT to the field of video makeup transfer, generating smooth and stable results.
However, the computational complexity of proposed SSCFT is quadratic to image size, the focus of our later work is to solve this problem.

%


\section{Acknowledgments}
This work was in part supported by NSFC (Grant No. 62176194) and the Major project of IoV, Technological Innovation Projects in Hubei Province (Grant No. 2020AAA001, 2019AAA024) and Sanya Science and Education Innovation Park of Wuhan University of Technology (Grant No. 2020KF0054), the National Natural Science Foundation of China under Grant 62101393 and the Open Project of Wuhan University of Technology Chongqing Research Institute(ZL2021--6).

\bibliographystyle{aaai21}
\bibliography{refs}

\clearpage
\section{Supplementary Materials}
\subsection{Generation of pseudo-paired data}
\begin{figure}[h]
\centering
\includegraphics[width=0.45\textwidth]{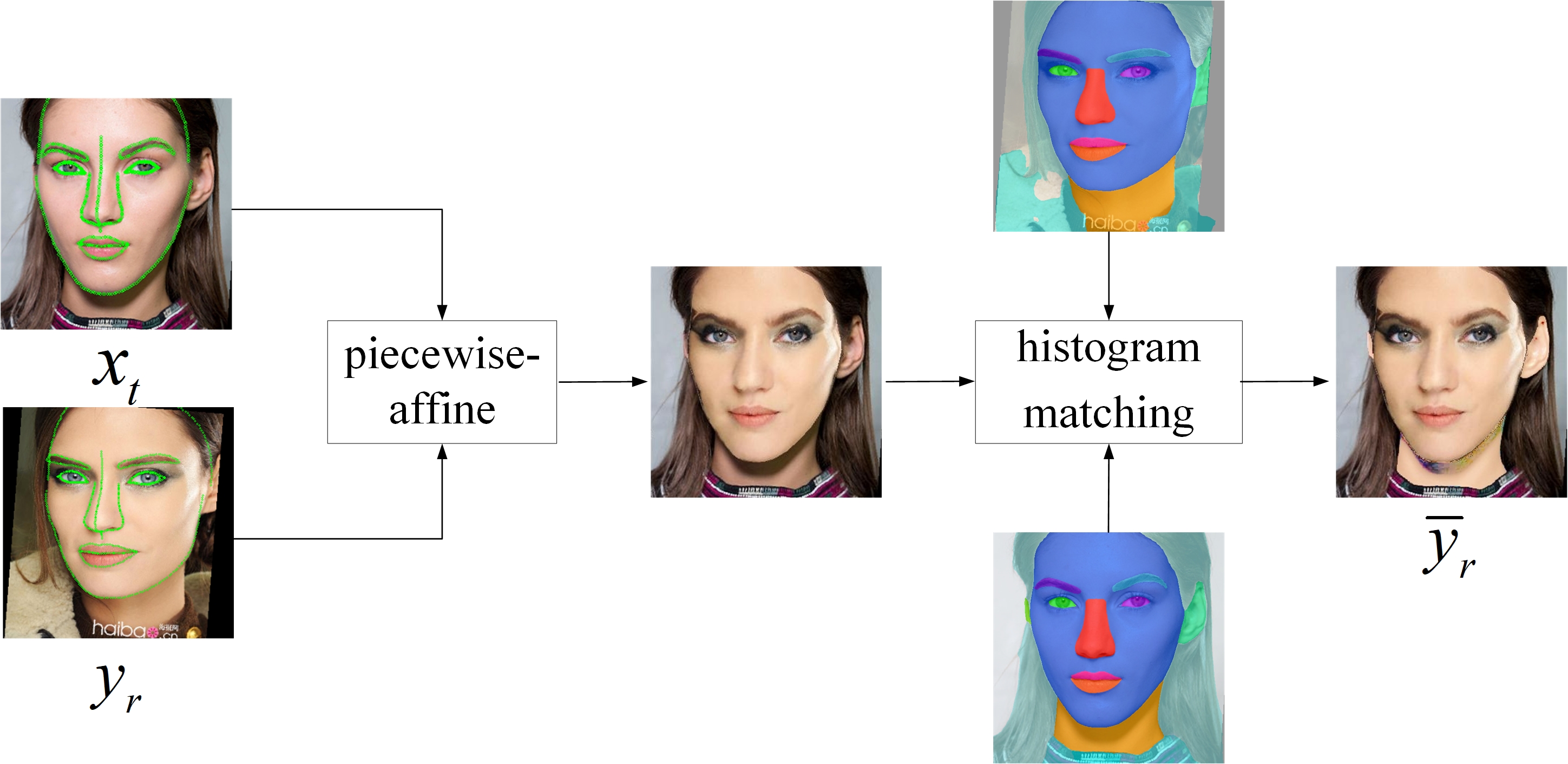} 
\caption{The process of generating pseudo-paired data.}
\label{img:groundtrue}
\end{figure}
Pseudo-paired data is used in the $L_{makeup}$, and the generation process is shown in Figure \ref{img:groundtrue}. Firstly, Face ++ API interface \footnote{https://www.faceplusplus.com.cn/dense-facial-landmarks/} is used to obtain the feature points of the input face image. Then, piecewise-affine transformation distorts the reference image into the target image according to the position of feature points. In this step, the inside of the eyes and the mouth remain unchanged, because these  two areas  need to be preserved during makeup transfer. Finally, histogram matching are applied to change the appearance of the ears and neck according to face parsing. Note that the warping results $\bar{y}_{r}$ sometimes possess artifacts, which can be fixed by SSAT in the generated results. For the dataset, we randomly selected 300 non-makeup images as target images and 300 makeup images as reference images from the Makeup Transfer dataset. Using the  proposed generation method, a total of 180,000 pairs of pseudo-paired data are generated for makeup transfer and removal. During the training, images are resized to $286\times286$, randomly cropped to $256\times256$, horizontally flipped with probability of 0.5, and randomly rotated from -30 degrees to +30 degrees for data augmentation.

\subsection{Network Structure}

For the network structure, the encoders consist of stacked convolution blocks, which contain convolution, Instance normalization, and ReLU activation layers.
In the decoder,  the SPADE is applyed to embed the makeup features into the content features, with each SPADE followed by a residual block.
For the discriminator, the network structure follows the multi-scale discriminator architecture.
The specific network structure parameters will be given in detail in the table below.
I, O, K, P, and S denote the number of input channels, the number of output channels, a kernel size, a padding size, and a stride size, respectively.
The network architecture of Encoder $E_{c}$ and $E_{m}$ has been shown in table \ref{table1}.
The network architecture of Encoder $E_{s}$ has been shown in table \ref{table2}
The network architecture of $FF$ module has been shown in table \ref{table3}
The network architecture of Encoder $Dec$ has been shown in table \ref{table4}
\begin{table}[h]
\centering
\begin{tabular}{c|c}
    \hline
    Layer & $E_{c}$ and $E_{m}$ \\
    \hline
    \hline
    L1 &  \makecell[c]{Conv(I:3,O:64,K:7,P:3,S:1),\\ Instance Normalization,\\ Leaky ReLU:0.2 } \\
    \hline
    L2 &  \makecell[c]{Conv(I:64,O:128,K:3,P:1,S:2),\\ Instance Normalization,\\ Leaky ReLU:0.2 } \\
    \hline
    L3 &  \makecell[c]{Conv(I:128,O:256,K:3,P:1,S:2),\\ Instance Normalization,\\ Leaky ReLU:0.2 } \\
    \hline
\end{tabular}
\caption{The network architecture of Encoder $E_{c}$ and $E_{m}$.}
\label{table1}
\end{table}

\begin{table}[h]
\centering
\begin{tabular}{c|c}
    \hline
    Layer & $E_{s}$ \\
    \hline
    \hline
    L1 &  \makecell[c]{Conv(I:18,O:32,K:7,P:3,S:1),\\ Instance Normalization,\\ Leaky ReLU:0.2 } \\
    \hline
    L2 &  \makecell[c]{Conv(I:32,O:64,K:3,P:1,S:2),\\ Instance Normalization,\\ Leaky ReLU:0.2 } \\
    \hline
    L3 &  \makecell[c]{Conv(I:64,O:128,K:3,P:1,S:2),\\ Instance Normalization,\\ Leaky ReLU:0.2 } \\
    \hline
\end{tabular}
\caption{The network architecture of Encoder $E_{s}$.}
\label{table2}
\end{table}

\begin{table}[h]
\centering
\begin{tabular}{c|c}
    \hline
    Layer & $FF$ \\
    \hline
    \hline
    L1 &  \makecell[c]{Conv(I:384,O:512,K:3,P:1,S:2),\\ Instance Normalization,\\ Leaky ReLU:0.2 } \\
    \hline
    L2 &  \makecell[c]{Conv(I:512,O:512,K:3,P:1,S:1),\\ Instance Normalization,\\ Leaky ReLU:0.2 } \\
    \hline
    L3 &  \makecell[c]{Connect [L1, L2, L3 of $E_{c}$,\\ L3 of $E_{s}$,\\ L1, L2 of $FF$]} \\
    \hline
\end{tabular}
\caption{The network architecture of $FF$ module.}
\label{table3}
\end{table}

\begin{table}[h]
\centering
\begin{tabular}{c|c}
    \hline
    Layer & $Dec$ \\
    \hline
    \hline
    L1 &  \makecell[c]{Upsample:2,\\ SPADE,\\ Resnet } \\
    \hline
    L2 &  \makecell[c]{Upsample:2,\\ SPADE,\\ Resnet } \\
    \hline
    L3 &  \makecell[c]{SPADE,\\ Resnet } \\
    \hline
    L4 &  \makecell[c]{Conv(I:64,O:3,K:7,P:3,S:1),\\ tanh } \\
    \hline
\end{tabular}
\caption{The network architecture of Encoder $Dec$.}
\label{table4}
\end{table}

\clearpage
\subsection{Makeup Style Interpolation}


Adjusting the shade of makeup style is an essential function of existing makeup applications. Due to the separation of makeup features, our method could generate  continuous makeup transfer results by interpolating makeup features, see  Figure \ref{img:supply_interpolation} . The formula is described as follows:

\begin{equation}\label{equ15}
  \hat{y}_{t}^{adjust}= Dec(x_{t}^{c}, (\hat{y}_{r}^{m_{1}} \cdot \alpha_{1} + \hat{y}_{r}^{m_{2}} \cdot \alpha_{2} ))
\end{equation}
where $\alpha_{1}+\alpha_{2}=1$. The closer $\alpha_{1}$ is to 1, the closer the resulting makeup style is to $\hat{y}_{r}^{m_{1}}$. The closer $\alpha_{2}$ is to 1, the closer the resulting makeup style is to $\hat{y}_{r}^{m_{2}}$.

\subsection{More Results}
\begin{figure}[h]
\centering
\includegraphics[width=0.45\textwidth]{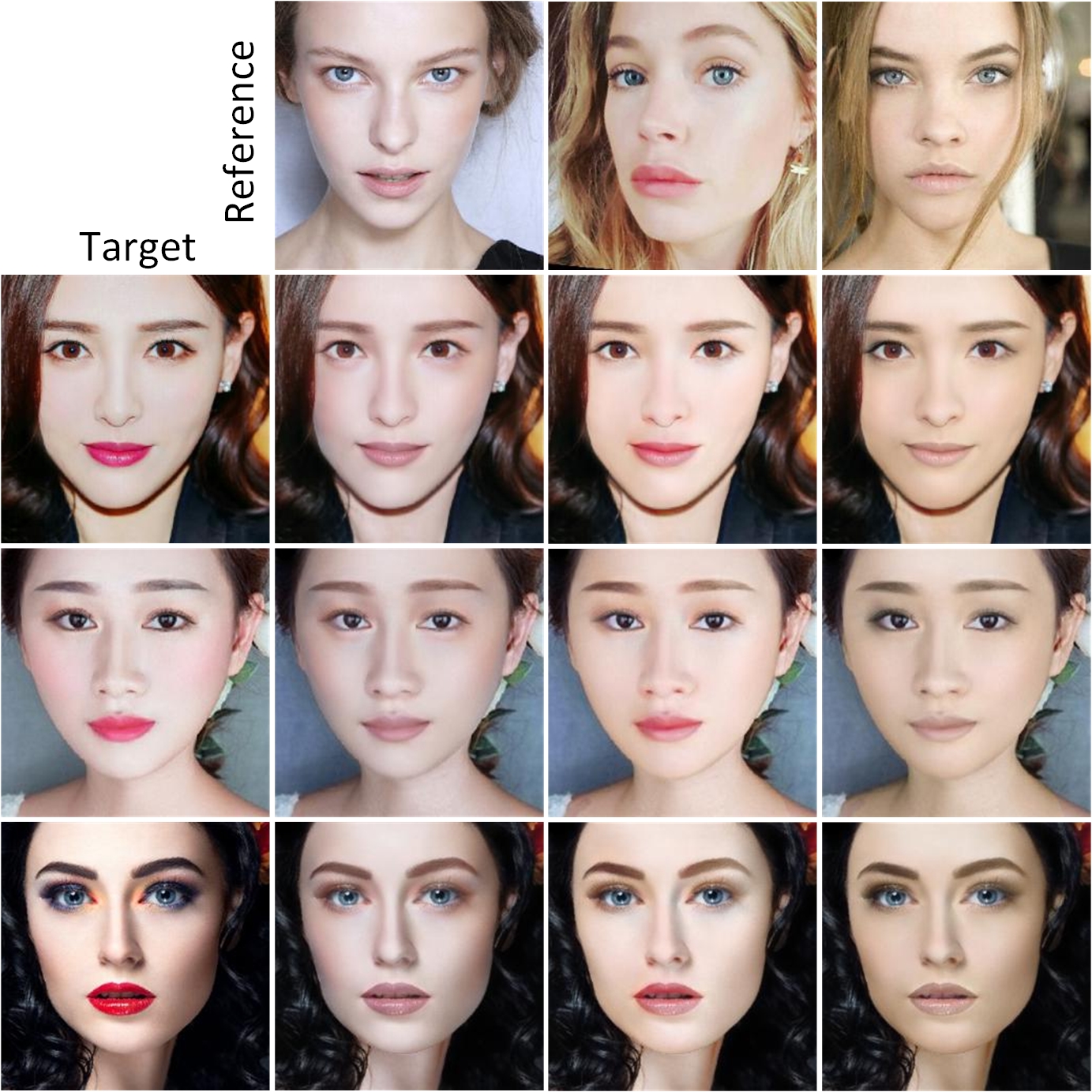} 
\caption{Our makeup removal results. The first column is three makeup target images, the first row is three non-makeup reference images,  the makeup removal results are displayed in the lower right corner.
}
\label{img:supply_removal}
\end{figure}

\begin{figure}[h]
\centering
\includegraphics[width=0.5\textwidth]{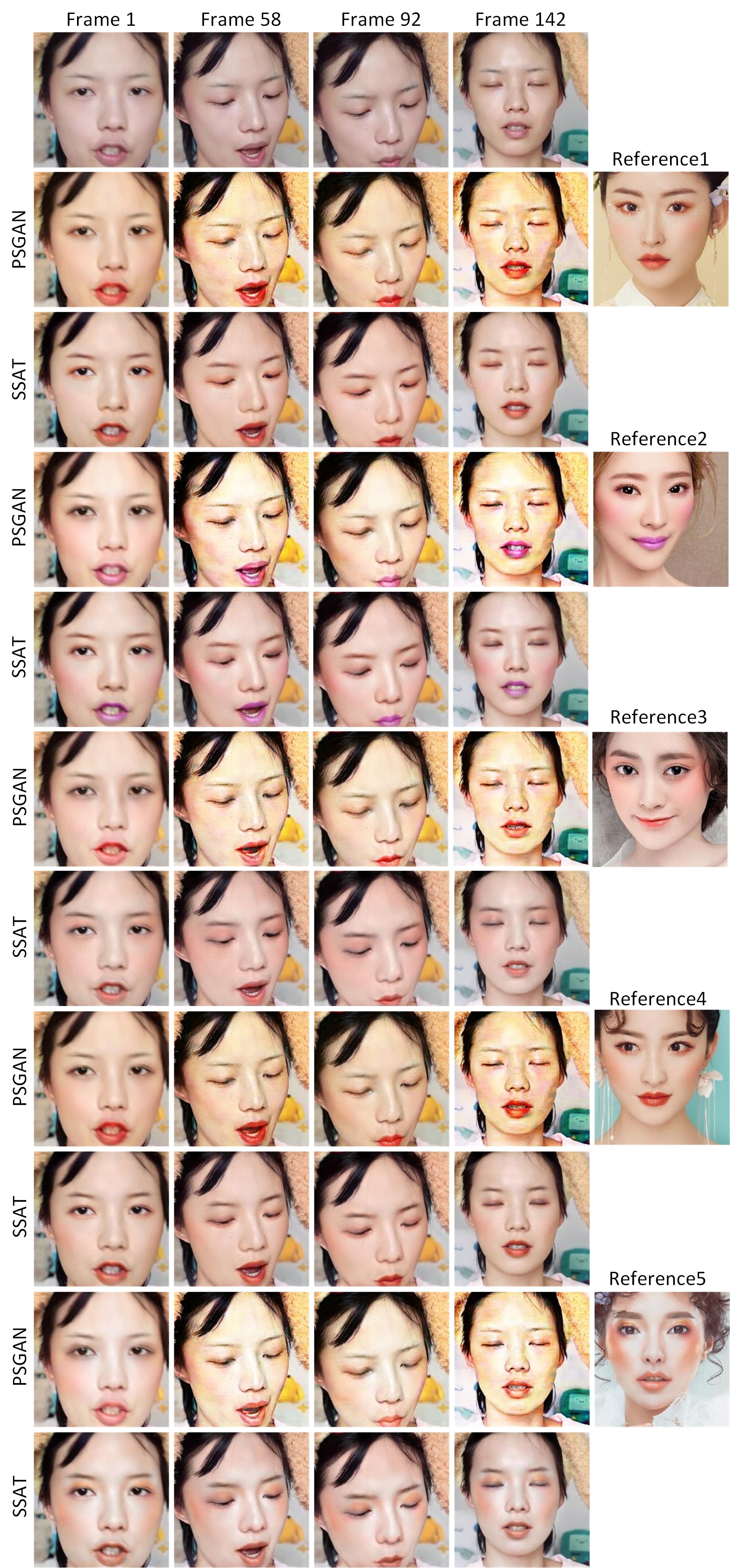} 
\caption{The video makeup transfer results. We also provide the full video results in the supplementary materials.
}
\label{img:supply_video}
\end{figure}

\begin{figure*}[t]
\centering
\includegraphics[width=1\textwidth]{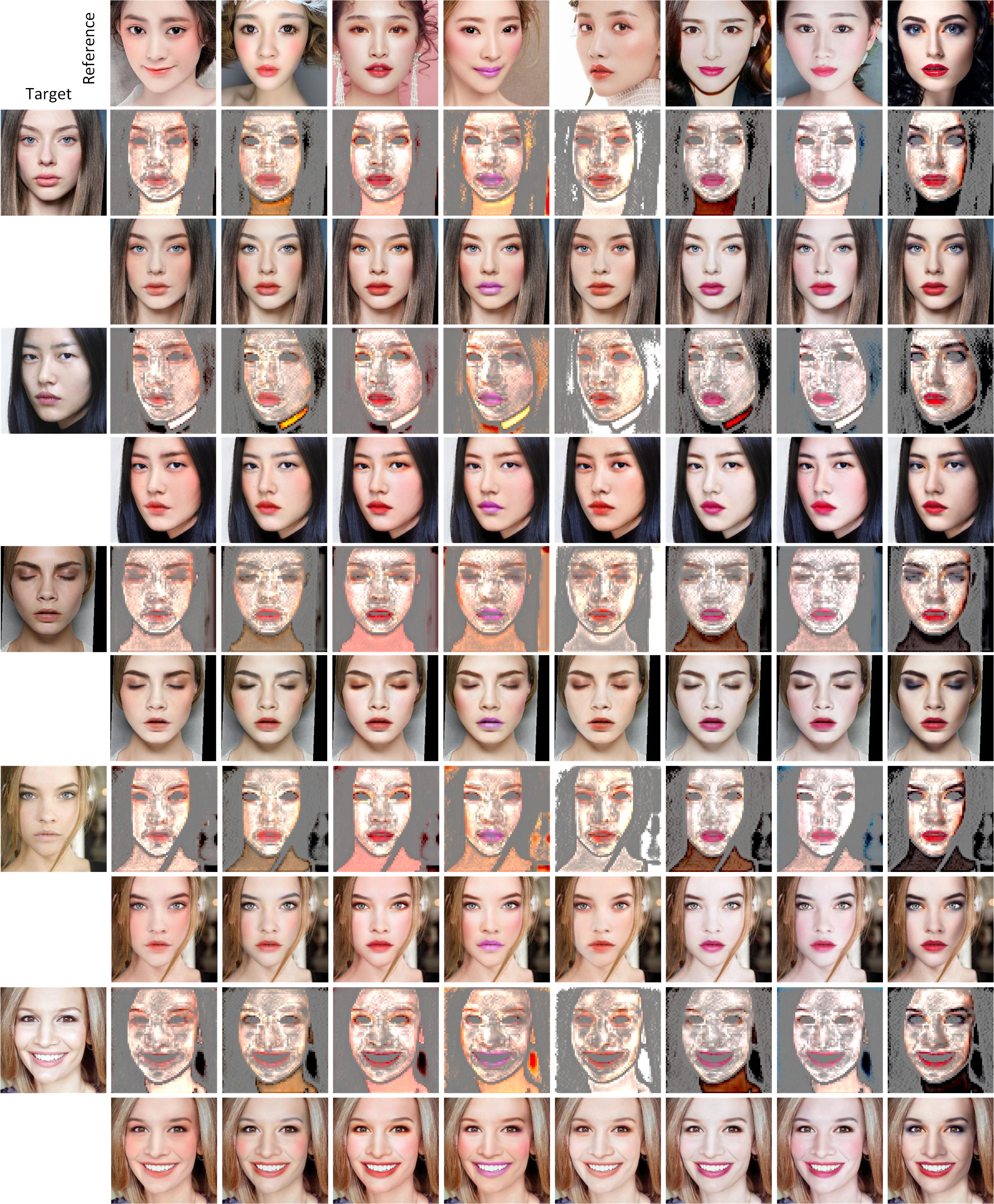} 
\caption{Our makeup transfer results. The first column is five target images, the first row is eight reference images, the semantic correspondence results and the makeup transfer results are displayed in the lower right corner. In the semantic correspondence results, the value of each spatial position of it is obtained by the different weighted sum of the reference image.
}
\label{img:supply_transfer}
\end{figure*}

\begin{figure*}[t]
\centering
\includegraphics[width=0.95\textwidth]{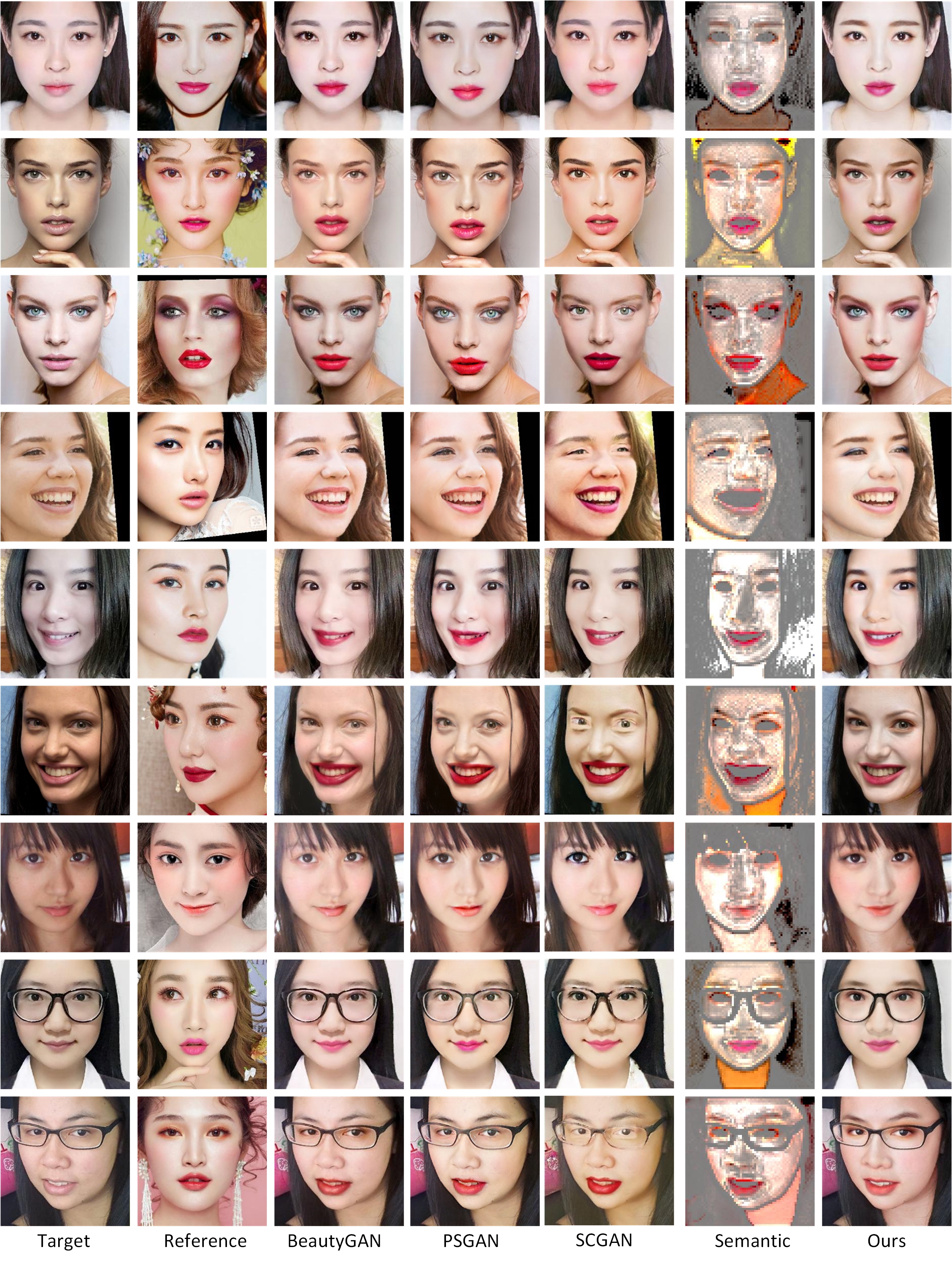} 
\caption{Comparison with state-of-the-art methods. The first three rows: frontal face, the middle three rows:  different expressions and poses, the last three rows: object occlusion (glasses or hair). From left to right, Target, Reference, BeautyGAN, PSGAN, SCGAN, Semantic, Ours. The value of each spatial position of semantic is obtained by the different weighted sum of the reference image according to semantic correspondence. In all cases, SSAT produces more accurate transfer results, especially for eye shadow and blush.
}
\label{img:supply_qualitative}
\end{figure*}

\begin{figure*}[t]
\centering
\includegraphics[width=1\textwidth]{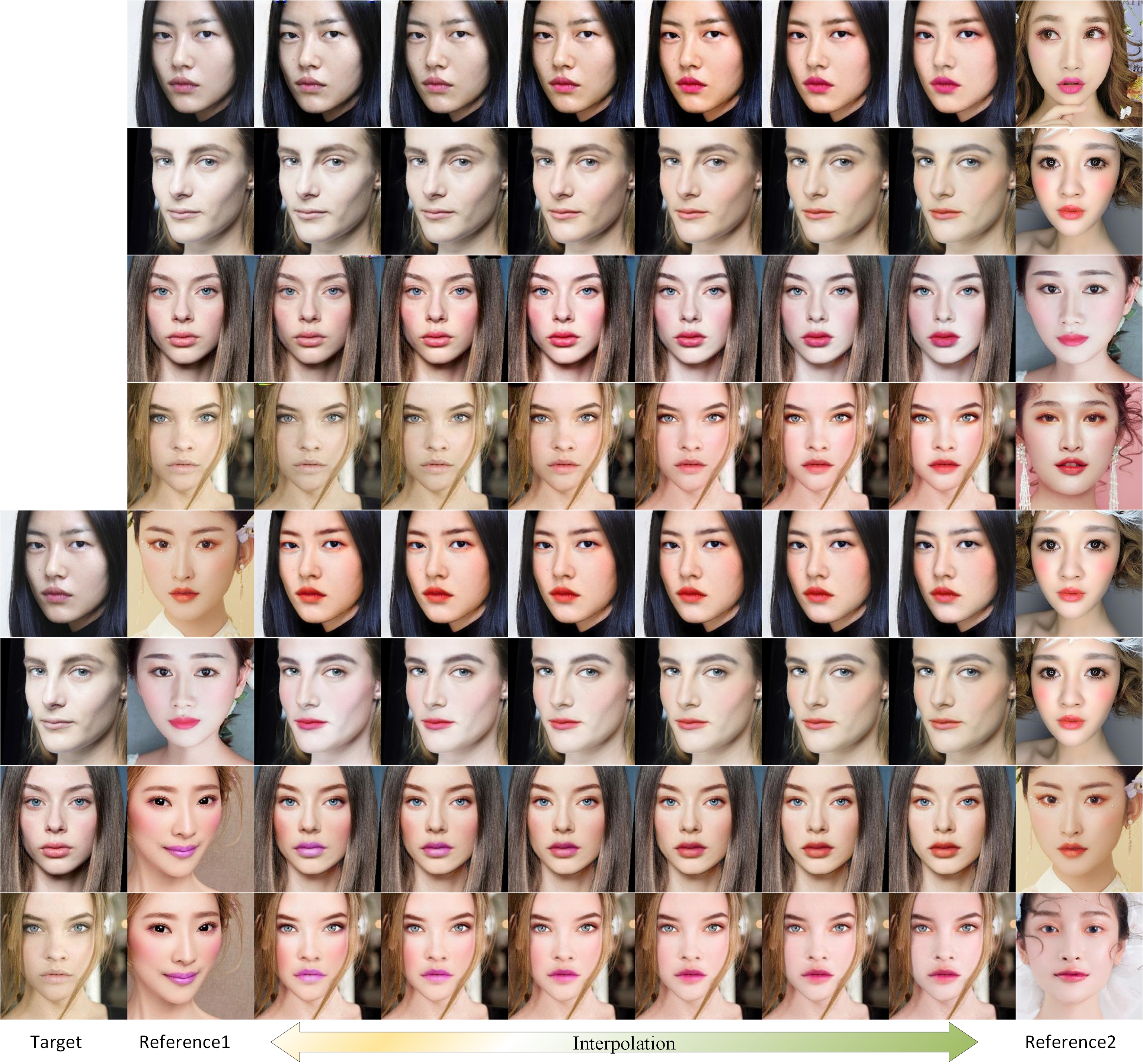} 
\caption{Results of makeup style interpolation. The first four rows have one reference image, and the last four rows have two. The makeup styles of the interpolated images continuously transfer from reference 1 to reference 2.
}
\label{img:supply_interpolation}
\end{figure*}

We present more results in additional material. The result of the makeup removal is shown in Figure \ref{img:supply_removal}. More video makeup transfer results are  shown in Figure \ref{img:supply_video}. More makeup transfer results is shown in Figure \ref{img:supply_transfer}. More qualitative comparisons are shown in Figure \ref{img:supply_qualitative}.  More results of makeup style interpolation are shown in Figure \ref{img:supply_interpolation}.

\clearpage


\end{document}